\documentstyle[aaai,amstex,times,epsf,url,fleqn]{article}

\renewcommand{\setminus}{-}

\newcommand{\nop}[1]{}

\newcommand{\bird}{\textit{bird}}
\newcommand{\fly}{\textit{fly}}
\newcommand{\winged}{\textit{metal\_wings}}
\newcommand{\penguin}{\textit{penguin}}
\newcommand{\easytosee}{\textit{easy\_to\_see}}
\newcommand{\havelegs}{\textit{legs}}

\newcommand{\yellow}{\textit{yellow}}
\newcommand{\magpie}{\textit{magpie}}
\newcommand{\chirps}{\textit{chirp}}

\newcommand{\nmableit}{\mbox{$\,|\hspace{-0.5em}\sim\,$}}
\newcommand{\nnmableit}{\mbox{$\,\|\hspace{-0.5em}\sim\,$}}

\newtheorem{remark}{Remark}
\newtheorem{theorem}{Theorem}
\newtheorem{corollary}[theorem]{Corollary}
\newtheorem{lemma}[theorem]{Lemma}

\newtheorem{proposition}[theorem]{Proposition}
\newtheorem{example}[theorem]{Example}

\newtheorem{definition}{Definition}
\newtheorem{algorithm}{Algorithm}

\newcommand{\be}{\begin{enumerate}}
\newcommand{\ee}{\end{enumerate}}
\newcommand{\bt}{\begin{tabular}}
\newcommand{\et}{\end{tabular}}
\newcommand{\bs}{\begin{theorem}}
\newcommand{\es}{\end{theorem}}
\newcommand{\bsw}[1]{\begin{theorem}[#1]}
\newcommand{\esw}{\end{theorem}}
\newcommand{\bc}{\begin{corollary}}
\newcommand{\ec}{\end{corollary}}
\newcommand{\bcw}[1]{\begin{corollary}[#1]}
\newcommand{\ecw}{\end{corollary}}
\newcommand{\ble}{\begin{lemma}}
\newcommand{\ele}{\end{lemma}}
\newcommand{\blew}[1]{\begin{lemma}[#1]}
\newcommand{\elew}{\end{lemma}}
\newcommand{\bp}{\begin{proposition}}
\newcommand{\ep}{\end{proposition}}
\newcommand{\bd}{\begin{definition}\rm}
\newcommand{\ed}{\end{definition}}
\newcommand{\bdw}[1]{\begin{definition}[#1]\rm}
\newcommand{\edw}{\end{definition}}
\newcommand{\ba}{\begin{algorithm}\rm}
\newcommand{\ea}{\end{algorithm}}
\newcommand{\baw}[1]{\begin{algorithm}[#1]\rm}
\newcommand{\eaw}{\end{algorithm}}
\newcommand{\bbs}{\begin{example}\rm}
\newcommand{\ebs}{\end{example}}
\newcommand{\bbsw}[1]{\begin{example}[#1]\rm}
\newcommand{\ebsw}{\end{example}}
\newcommand{\bb}{\begin{remark}\rm}
\newcommand{\eb}{\end{remark}}
\newcommand{\beq}{\begin{eqnarray*}}
\newcommand{\eeq}{\end{eqnarray*}}
\newcommand{\baq}{\begin{array}}
\newcommand{\eaq}{\end{array}}

\newcommand{\ins}{\,{\in}\,}

\newcommand{\gts}{\,{>}\,}

\newcommand{\les}{\,{\le}\,}
\newcommand{\lts}{\,{<}\,}
\newcommand{\eqs}{\,{=}\,}
\newcommand{\modelss}{\,{\models}\,}

\newcommand{\cups}{\,{\cup}\,}
\newcommand{\wedges}{\,{\wedge}\,}
\newcommand{\subseteqs}{\,{\subseteq}\,}
\newcommand{\supseteqs}{\,{\supseteq}\,}

\def\cl#1{\gamma(#1)}
\def\ras{\,{\ra}\,}
\def\Ras{\,{\Rightarrow}\,}

\def\tmodels{\models_{\mbox{\footnotesize\it tight}}}

\def\tzmodels{\models^{z}_{\mbox{\footnotesize\it tight}}}

\def\tlmodels{\models^{\it lex}_{\mbox{\footnotesize\it tight}}}
\def\tcmodels{\models^{\it ce}_{\mbox{\footnotesize\it tight}}}

\def\zmodels{\models^{z}}

\def\lmodels{\models^{\it lex}}
\def\cmodels{\models^{\it ce}}

\def\oazmodels{\nmableit^{z}}
\def\oasmodels{\nmableit^{s}}
\def\oalmodels{\nmableit^{\it lex}}
\def\oacmodels{\nmableit^{\it ce}}
\def\azmodels{\nnmableit^{z}}
\def\asmodels{\nnmableit^{s}}
\def\atmodels{\nnmableit^{\mbox{\!\tiny 1}}}
\def\aomodels{\nnmableit^{\mbox{\!\tiny 0}}}
\def\almodels{\nnmableit^{\it lex}}
\def\acmodels{\nnmableit^{\it ce}}
\def\tazmodels{\nnmableit^{z}_{\!\mbox{\footnotesize\it tight}}\,}
\def\tasmodels{\nnmableit^{s}_{\!\mbox{\footnotesize\it tight}}\,}
\def\talmodels{\nnmableit^{\it lex}_{\!\mbox{\footnotesize\it tight}}\,}
\def\tacmodels{\nnmableit^{\it ce}_{\!\mbox{\footnotesize\it tight}}\,}
\def\taomodels{\nnmableit^{\mbox{\!\tiny 0}}_{\!\mbox{\footnotesize\it tight}}\,}
\def\tatmodels{\nnmableit^{\mbox{\!\tiny 1}}_{\!\mbox{\footnotesize\it tight}}\,}

\def\tzmodels{\tazmodels}
\def\tlmodels{\talmodels}
\def\tcmodels{\tacmodels}
\def\zmodels{\azmodels}
\def\lmodels{\almodels}
\def\cmodels{\acmodels}

\def\cF{{\mathcal{F}}}

\def\cI{{\mathcal{I}}}

\def\Pr{{\it Pr}}

\def\cll#1#2#3{(#1|#2)[#3]}
\def\spcl#1#2#3{(#1\,|\,#2)[#3]}
\def\dspcl#1#2#3{(#1\,\|\,#2)[#3]}

\def\KB{{\it K\!B}}
\def\BG{{\it T}}

\def\isP{\cI_{\Phi}}

\def\ra{\rightarrow}

\nocopyright

\title{Probabilistic Default Reasoning\\ 
with Conditional Constraints}

\author{Thomas Lukasiewicz\\[1ex]
Institut und Ludwig Wittgenstein Labor f\"ur Informationssysteme, TU Wien\\
Favoritenstra\ss{}e\ 9-11, A-1040 Vienna, Austria\\
lukasiewicz@@kr.tuwien.ac.at}

\begin{document}

\maketitle

\begin{abstract}
We propose a combination of probabilistic reasoning from
conditional constraints with approaches to default reasoning 
from conditional knowledge bases. 
In detail, we generalize the notions of Pearl's entailment in system $Z$, 
Lehmann's lexicographic entailment, and Geffner's conditional entailment 
to conditional constraints.
We give some examples that show that the new notions of $z$-, lexicographic, 
and conditional entailment have similar properties like their classical counterparts. 
Moreover, we show that the new notions of $z$-, lexicographic, 
and conditional entailment are proper generalizations of both their classical counterparts
and the classical notion of logical entailment for conditional constraints.  
\end{abstract}

\setcounter{page}{1}

\section{Introduction}

In this paper, we elaborate a combination of
probabilistic reasoning from conditional constraints
with approaches to default reasoning from conditional knowledge bases.
As a main result, this combination provides new notions of entailment for conditional constraints, 
which respect the ideas of classical default reasoning from conditional knowledge bases, 
and which are generally much stronger than the classical notion 
of logical entailment based on conditioning. Moreover,
the results of this paper can also be applied for 
handling inconsistencies in probabilistic knowledge bases. 

Informally, the ideas behind this paper can be described as follows. 
Assume that we have the following knowledge at hand: ``all penguins are birds'' (G1),
``between 90 and 95\% of all birds fly'' (G2), 
and ``at most 5\% of all penguins fly'' (G3). 
Moreover, assume a first scenario in which 
``Tweety is a bird'' (E1)
and second one in which ``Tweety is a penguin'' (E2).
What do we conclude about Tweety's ability to fly?

A closer look at this example shows that the statements G1--G3
describe statistical knowledge (or objective knowledge), 
while E1 and E2 express degrees of belief (or subjective knowledge).   
One way of handling such combinations of statistical knowledge 
and degrees of belief is {\em reference class reasoning}, 
which goes back to Reichenbach (1949) \nocite{Rei49}
and was further refined by Kyburg (1974; 1983) \nocite{Kyb74,Kyb83} and Pollock (1990). \nocite{Pollock90}

Another related field is {\em default reasoning from conditional knowledge bases}, 
where we have generic statements of the form ``all penguins are birds'', 
``generally, all birds fly'', and ``generally, no penguin flies'' in 
addition to some concrete evidence as E1 and E2.  
The literature contains several different
approaches to default reasoning
and extensive work on the desired properties.
The core of these properties are the rationality postulates 
proposed by Kraus {\em et al.}\ (1990). \nocite{KLM90}
These rationality postulates constitute 
a sound and complete axiom system for several classical model-theoretic entailment
relations under uncertainty measures on worlds. 
In detail, they characterize classical model-theoretic entailment
under preferential structures (Shoham 1987; Kraus {\em et al.}\ 1990), \nocite{Shoham87,KLM90} 
infinitesimal probabilities (Adams 1975; Pearl 1989), \nocite{Ada75,Pearl89} 
possibility measures (Dubois \& Prade 1991), \nocite{DP91} 
and world rankings (Spohn 1988; Goldszmidt \& Pearl 1992). \nocite{Spohn88,GP92}. 
They also characterize 
an entailment relation based on conditional objects (Dubois \& Prade 1994). \nocite{DP94}  
A survey of all these relationships is given in (Benferhat {\em et al.}\ 1997). \nocite{BDH97}
Recently, Friedman and Halpern (2000) \nocite{FH99}
showed that many approaches describe to the same notion of inference, 
since they are all expressible as plausibility measures.

Mainly to solve problems with irrelevant information, the notion of
rational closure as a more adventurous notion of entailment has been
introduced by Lehmann (Lehmann 1989; Lehmann \& Magidor 1992). \nocite{Lehmann89,lehmann_d-magidor:1992b1} 
This notion of entailment is equivalent to entailment in system $Z$ by
Pearl (1990), \nocite{Pearl90} to the least specific possibility
entailment by Benferhat {\em et al.}\ (1992), \nocite{BDP92} and to a conditional
(modal) logic-based entailment by Lamarre (1992)\nocite{Lamarre92}.
Finally, mainly in order to solve problems with property inheritance from
classes to exceptional subclasses, the maximum entropy approach to
default entailment was proposed by Goldszmidt {\em et al.}\ (1993);\nocite{GPM93} 
the notion of lexicographic entailment was
introduced by Lehmann (1995)\nocite{Persp:94} and Benferhat {\em et al.}\ (1993);
\nocite{BCDLP:lexi} the notion of conditional entailment was proposed
by Geffner (Geffner 1992; Geffner \& Pearl 1992); \nocite{geffner:1992c,GeffnerPearl92} and an infinitesimal
belief function approach was suggested by Benferhat {\em et al.}\ (1995).\nocite{benf-etal-98}

Coming back to our introductory example, we realize that
G1--G3 and E1--E2 represent {\em interval restrictions for conditional probabilities}, 
also called {\em conditional constraints} (Lukasiewicz 1999b). \nocite{Luk99a} The literature 
contains extensive work on reasoning about conditional constraints 
(Dubois \& Prade 1988; Dubois {\em et al.}\ 1990; 1993; 
Amarger {\em et al.}\ 1991; Jaumard {\em et al.}\ 1991; Th\"{o}ne {\em et al.}\ 1992;
Frisch \& Haddawy 1994; Heinsohn 1994; 
Luo {\em et al.}\ 1996; Lukasiewicz 1999a; 1999b)
\nocite{DP88,DPT90,ADP91a,jahapo91,TGK92,DPGM93,FH94+,Hein94,LYLWP96,Luk99a,Luk99b} 
and their generalizations, 
for example, to probabilistic logic programs (Lukasiewicz 1998). \nocite{Luk98c}
 
Now, the main idea of this paper is to use techniques for default reasoning from 
conditional knowledge bases in order to perform probabilistic reasoning from 
statistical knowledge and degrees of beliefs. More precisely, 
we extend the notions of entailment in system $Z$, Lehmann's lexicographic entailment, 
and Geffner's conditional entailment to the framework of 
conditional constraints. 

Informally, in our introductory example, the statements 
G2 and G3 are interpreted as ``generally, a bird flies with a probability between 0.9 and 0.95'' 
(G2$^{\star}$) and ``generally, a penguin flies with a probability of 
at most \mbox{0.05'' (G3$^{\star}$)}, respectively. 
In the first scenario, we then simply use the whole probabilistic 
knowledge $\{\text{G1},\text{G2}^{\star},\text{G3}^{\star},\text{E1}\}$ to conclude
under classical logical entailment that
``Tweety flies with a probability between 0.9 and 0.95''. 
In the second scenario, it turns out that the whole probabilistic 
knowledge $\{\text{G1},\text{G2}^{\star},\text{G3}^{\star},\text{E2}\}$
is unsatisfiable.  More precisely, $\{\text{G1},\text{G2}^{\star},\text{G3}^{\star}\}$ is inconsistent 
in the context of a penguin. In fact, the main problem is
that G2$^{\star}$ should not be applied anymore to penguins. 
\mbox{That is}, we can easily resolve the inconsistency by removing G2$^{\star}$, 
and then conclude from $\{\text{G1},\text{G3}^{\star},\text{E2}\}$ under classical logical entailment
that ``Tweety flies with a probability of \mbox{at most 0.05}''. 

Hence, the results of this paper can also be used for handling inconsistencies in
probabilistic knowledge bases. More precisely, the new notions of nonmonotonic entailment 
coincide with the classical notion of logical entailment as far as {\em satisfiable}
sets of conditional constraints are concerned.
Furthermore, they allow desirable conclusions from certain 
kinds of {\em unsatisfiable} sets of conditional constraints. 

We remark that this inconsistency handling is guided by the principles of default 
reasoning from conditional knowledge bases. It is     
thus based on a natural preference relation
on conditional constraints, and not on the assumption
that all conditional constraints are equally weighted 
(as, for example, in the work by Jaumard {\em et al.}\ (1991)).
\nocite{jahapo91}

The work closest in spirit to this paper is perhaps the 
one by Bacchus {\em et al.}\ (1996), \nocite{BGHK96} which suggests to use 
the {\em random worlds method} (Grove {\em et al.}\ 1994) \nocite{GHK94} to induce degrees of beliefs
from quite rich statistical knowledge bases. However, differently
from (Bacchus {\em et al.}\ 1996), \nocite{BGHK96} we do not make use of a strong principle such as the 
random worlds method (which is closely related to probabilistic reasoning 
under maximum entropy). Moreover, we restrict 
our considerations to the propositional setting. 

The main contributions of this paper are as follows: 
\be
\item[$\bullet$] We illustrate that 
the classical notion of logical entailment for conditional constraints
is not very well-suited for default reasoning with conditional constraints.
\item[$\bullet$] We introduce the notions of $z$-entailment, lexicographic entailment, 
and conditional entailment for conditional constraints, which are 
a combination of the classical notions of entailment in system $Z$ (Pearl 1990), \nocite{Pearl90} 
Lehmann's lexicographic entailment (Lehmann 1995), \nocite{Persp:94} and Geffner's conditional entailment 
(Geffner 1992; Geffner \& Pearl 1992),\nocite{geffner:1992c,GeffnerPearl92} respectively, with the
classical notion of logical entailment for conditional constraints.
\item[$\bullet$] We give some examples that analyze the nonmonotonic properties of 
the new notions of entailment for default reasoning with conditional constraints.
It turns out that the new notions of $z$-entailment, lexicographic entailment, 
and conditional entailment have similar properties like their classical counterparts. 
\item[$\bullet$] We show that the new notions of $z$-entailment, lexicographic entailment, 
and conditional entailment for conditional constraints properly extend the classical notions of entailment in system $Z$, 
lexicographic entailment, and conditional entailment, respectively.
\item[$\bullet$] We show that the new notions of $z$-entailment, lexicographic entailment, 
and conditional entailment for conditional constraints properly extend the classical 
notion of logical entailment for conditional constraints.
\ee

Note that all proofs are given in (Lukasiewicz 2000). \nocite{Lukas00-1} 

\section{Preliminaries}\label{SEC-1}
We now introduce some necessary technical background. 

We assume a finite nonempty set of {\em basic propositions} (or {\em atoms}) $\Phi$. 
We use  $\bot$ and $\top$ to denote the propositional constants {\em false} and {\em true}, 
respectively. The set of {\em classical formulas} is the closure of $\Phi\cups\{\bot,\top\}$ 
under the Boolean operations $\neg$ and $\wedge$. 
A {\em strict conditional constraint} is an expression
$(\psi|\phi)[l,u]$ with real 
numbers $l,u\ins [0,1]$ and classical formulas $\psi$ and $\phi$.
A {\em defeasible conditional constraint} (or {\em default}) is an expression
$(\psi\|\phi)[l,u]$ with real 
numbers $l,u\ins [0,1]$ and classical formulas $\psi$ and $\phi$.
A {\em conditional constraint} is a strict or defeasible conditional constraint. 
The set of {\em strict probabilistic formulas} (resp., {\em probabilistic formulas}) is the 
closure of the set of all strict conditional constraints (resp., conditional constraints) 
under the Boolean operations $\neg$ and $\wedge$. We use $(F\,{\vee}\,G)$, 
$(F\,{\Rightarrow}\,G)$, and $(F\,{\Leftrightarrow}\,G)$
to abbreviate $\neg(\neg F\wedges \neg G)$, $\neg (F\wedges\neg G)$, and 
$(\neg (\neg F\wedges G))\wedges(\neg (F\wedges\neg G))$, respectively, and adopt the usual
conventions to eliminate parentheses.

A {\em probabilistic default theory} 
is a pair $\BG\eqs(P,D)$, where $P$ is a finite set of strict conditional constraints and 
$D$ is a finite set of defeasible conditional constraints. 
A {\em probabilistic knowledge base} $\KB$ is a strict probabilistic formula.   
Informally, default theories represent strict and defeasible generic knowledge,
while probabilistic knowledge bases express some concrete evidence.  

A {\em possible world} is a truth assignment $I\colon \Phi
\ra \{\mbox{\bf true}$, $\mbox{\bf false}\}$, which is extended to classical
formulas as usual. We use ${\cal I}_{\Phi}$ to denote the set of all
possible worlds for $\Phi$. A possible world $I$ {\em satisfies} a classical
\mbox{formula $\phi$}, or $I$ is a {\em model} \mbox{of $\phi$}, denoted $I\models
\phi$, iff $I(\phi)=\mbox{\bf true}$.

A {\em probabilistic interpretation} $\Pr$ is a probability function on ${\cal I}_{\Phi}$ (that is, 
a mapping $\Pr\colon{\cal I}_{\Phi}\ras[0,1]$ such that all $\Pr(I)$ with $I\ins{\cal I}_{\Phi}$ sum up to 1).
The {\em probability} of a classical \mbox{formula $\phi$} 
in the probabilistic interpretation $\Pr$, denoted $\Pr(\phi)$, is defined as follows:
\[\baq{l@{\ \ }l@{\ \ }l}
\Pr(\phi) & = & \sum\limits_{I\in\,{\cal I}_{\Phi},\,I\,\models \phi} \Pr(I)\,.
\eaq
\]
For classical formulas $\phi$ and $\psi$ with $\Pr(\phi)\gts 0$, we use $\Pr(\psi|\phi)$ 
to abbreviate $\Pr(\psi\wedge\phi)\,/\,\Pr(\phi)$.
The {\em truth} of probabilistic formulas $F$ in a probabilistic interpretation $\Pr$, 
denoted $\Pr\models F$, is inductively defined as follows:
\begin{itemize}
\item $\Pr\models \cll{\psi}{\phi}{l,u}$ \ iff \ $\Pr(\phi)\eqs 0$ or $\Pr(\psi|\phi)\ins[l,u]$.
\item $\Pr\models (\psi\|\phi)[l,u]$ \ iff \ $\Pr(\phi)\eqs 0$ or $\Pr(\psi|\phi)\ins[l,u]$.
\item $\Pr\models \neg F$ \ iff \ not $\Pr\models F$.
\item $\Pr\models (F\wedge G)$ \ iff \ $\Pr\models F$ and $\Pr\models G$.
\end{itemize}

We remark that there is no difference between strict 
and defeasible conditional constraints as far as the 
notion of truth in probabilistic interpretations is concerned.  

A probabilistic interpretation $\Pr$ {\em satisfies} a probabilistic \mbox{formula $F$}, 
or $\Pr$ is a {\em model} \mbox{of $F$}, iff $\Pr\modelss F$.
\mbox{$\Pr$ {\em satisfies}} a set of probabilistic formulas
$\cF$, or $\Pr$ is a {\em model} of $\cF$, denoted 
$\Pr\models \cF$, iff $\Pr$ is a model of all $F\ins\cF$. 
We say $\cF$ is {\em satisfiable} iff a model of $\cF$ exists. 

We next define the notion of logical entailment as follows.
A strict probabilistic formula $F$ is a {\em logical consequence} of a set of probabilistic formulas $\cF$, 
denoted $\cF\modelss F$, iff each model of $\cF$ is also a model of $F$. 
A strict conditional constraint $(\psi|\phi)[l,u]$ is a {\em tight logical consequence} 
\mbox{of $\cF$}, denoted $\cF\,{\tmodels}\,$$(\psi|\phi)[l,u]$,
iff $l$ (resp., $u$) is the infimum (resp., supremum) of
$\Pr(\psi|\phi)$ 
subject to all models $\Pr$ of $\cF$ with $\Pr(\phi)\gts 0$ 
(note that we canonically define $l\eqs 1$ and $u\eqs 0$, 
when $\cF\modelss$$(\phi|\top)[0,0]$).

We remark that every notion of entailment for conditional 
constraints is associated with a notion of consequence and 
a notion of tight consequence. Informally, the notion of consequence
describes entailed intervals, while the notion of tight consequence
characterizes the tightest entailed interval. That is,  
if $(\psi|\phi)[l,u]$ is a tight consequence 
of $\cF$, then $[l',u']\supseteqs[l,u]$ for all consequences 
$(\psi|\phi)[l',u']$ of $\cF$.

\begin{table*}[htb]
\renewcommand{\arraystretch}{1.3}
\caption{\label{EX-TAB-1}Examples of 0- and 1-entailed tight intervals.}

\medskip 
\begin{center}
\bt{|c||c|c|c||c|c|}\hline
 & $\BG$ & $\KB$ & $(\psi|\phi)$ & $\taomodels$ & $\tatmodels$  \\\hline\hline
(1) & $\BG_1$ & $\spcl{\bird}{\top}{1,1}$ & $(\havelegs|\top)$ & $[.95,1]$ & $[.95,1]$ \\
(2) & $\BG_1$ & $\spcl{\bird\wedge\yellow}{\top}{1,1}$ & $(\havelegs|\top)$ & $\boldsymbol{[0,1]}$ & $[.95,1]$ \\
(3) & $\BG_1$ & $\spcl{\penguin}{\top}{1,1}$ & $(\havelegs|\top)$ & $\boldsymbol{[0,1]}$ & $[.95,1]$ \\\hline
(4) & $\BG_2$ & $\spcl{\bird}{\top}{1,1}$ & $(\havelegs|\top)$ & $[.95,1]$ & $[.95,1]$ \\
(5) & $\BG_2$ & $\spcl{\penguin}{\top}{1,1}$ & $(\havelegs|\top)$ & $\boldsymbol{[0,1]}$ & $\boldsymbol{[1,0]}$ \\
(6) & $\BG_2$ & $\spcl{\bird}{\top}{1,1}$ & $(\fly|\top)$ & $[.9,.95]$ & $[.9,.95]$ \\
(7) & $\BG_2$ & $\spcl{\penguin}{\top}{1,1}$ & $(\fly|\top)$ & $[0,.05]$ & $\boldsymbol{[1,0]}$ \\\hline
(8) & $\BG_3$ & $\spcl{\penguin\wedge \yellow}{\top}{1,1}$ & $(\easytosee|\top)$ & $\boldsymbol{[0,1]}$ & $\boldsymbol{[1,0]}$\\\hline
(9) & $\BG_4$ & $\spcl{\magpie}{\top}{1,1}$ & $(\chirps|\top)$ & $[.7,.8]$  & $[.7,.8]$\\\hline
(10) & $\BG_5$ & $\spcl{\penguin\wedge \winged}{\top}{1,1}$ & $(\fly|\top)$ & $[0,1]$  & $\boldsymbol{[1,0]}$ \\\hline
(11) & $\BG_2$ & $\spcl{\bird}{\top}{.9,1}\wedge\spcl{\penguin}{\top}{.1,1}$ & $(\fly|\top)$ &{\bf undefined} & $[.86,.91]$ \\
(12) & $\BG_2$ & $\spcl{\bird}{\top}{.9,1}\wedge\spcl{\penguin}{\top}{.9,1}$ & $(\fly|\top)$ &{\bf undefined} & $\boldsymbol{[1,0]}$ \\\hline
\et
\end{center}
\renewcommand{\arraystretch}{1}
\end{table*}

\section{Motivating Examples}\label{SEC-2}

What should a probabilistic knowledge base entail
under a probabilistic default theory? To get a rough idea on
the reply to this question, we now introduce two natural
notions of entailment and analyze their properties. \mbox{It will} turn out 
that neither of these two notions is fully adequate  for probabilistic 
default reasoning with conditional constraints.  

In the sequel, let $\BG\eqs(P,D)$ be a probabilistic default theory. 
We first define the notion of 0-entailment, which applies to probabilistic 
knowledge bases of the form $\KB\eqs(\varepsilon|\top)[1,1]$. 
In detail, a strict conditional constraint $(\psi|\phi)[l,u]$ is a {\em $0$-consequence} of $\KB$,
denoted $\KB\aomodels (\psi|\phi)[l,u]$, iff $P\cups D\models (\psi|\phi\wedges\varepsilon)[l,u]$.
\mbox{It is a {\em tight}} \mbox{\em $0$-consequence} of $\KB$,
denoted $\KB\taomodels(\psi|\phi)[l,u]$, iff 
$P\cups D\tmodels(\psi|\phi\wedges\varepsilon)[l,u]$.
Informally, we use the concrete evidence in $\KB$ to 
fix our ``point of interest'' and  the generic knowledge 
in $\BG$ to draw the requested conclusion.   
That is, we perform classical conditioning. 

We next define the notion of 1-entailment, which  applies to all probabilistic 
knowledge bases $\KB$. 
\mbox{A strict} probabilistic formula $F$ is a {\em $1$-consequence} of $\KB$,
denoted $\KB\atmodels F$, iff $P\cup D\cup \{\KB\}\models F$.
A strict conditional constraint $(\psi|\phi)[l,u]$ is a {\em tight $1$-consequence} of $\KB$,
denoted $\KB\tatmodels(\psi|\phi)[l,u]$, iff 
$P{\cup} D{\cup} \{\KB\}\,{\tmodels}\,(\psi|\phi)[l,u]$.
Informally, we draw our conclusion from 
the union of the concrete evidence in $\KB$ and the generic 
knowledge in $\BG$.   

We now analyze the properties of these two notions of entailment.
Our first example concentrates on the aspects of {\em ignoring irrelevant information}
and {\em property inheritance}.
\bbs\label{EX-1}
The knowledge ``all penguins are birds''
and \mbox{``at least} 95\% of all birds have legs'' 
can be expressed by the following probabilistic default theory $\BG_1\eqs(P_1,D_1)$:
\[\baq{l@{\ \ }l@{\ \ }l}
P_1 & = &\{\spcl{\bird}{\penguin}{1,1}\},\\[0.8ex]
D_1 & = &\{\dspcl{\havelegs}{\bird}{.95,1}\}\,. 
\eaq\]

Now, $\BG_1$ should entail that ``generally,
birds have legs with a probability of at least 0.95'' (that is, e.g., 
if we know that Tweety is a bird, and we do not have any other knowledge, 
then we should conclude that the probability of
Tweety having legs is at least 0.95). Indeed, this conclusion
is drawn under both 0- and 1-entailment (see
item (1) in Table \ref{EX-TAB-1}).  

Moreover, $\BG_1$ should entail that ``generally, yellow 
birds have legs with a probability of at least 0.95''
(as the property ``yellow'' is not mentioned at all in $\BG_1$ and thus {\em irrelevant}),
and that ``generally, penguins have legs with a probability of at 
least 0.95'' (as 
the set of all penguins is a {\em nonexceptional subclass}
of the set of all birds, and thus penguins should {\em inherit} all properties of birds). 
However, while 1-entailment still allows the desired conclusions, 
0-entailment just yields the interval $[0,1]$ (see items (2)--(3) in Table \ref{EX-TAB-1}). $\Box$
\ebs

We next concentrate on the principle of {\em specificity} and
the problem of {\em inheritance blocking}. 
\bbs\label{EX-2}
Let us consider the following probabilistic default theory $\BG_2\eqs(P_2,D_2)$:
\[\baq{l@{\ \ }l@{\ \ }l}
P_2 & = &\{\spcl{\bird}{\penguin}{1,1}\},\\[0.8ex]
D_2 & = &\{\dspcl{\havelegs}{\bird}{.95,1},\,
\dspcl{\fly}{\bird}{.9,.95},\,\\[0.8ex]
&&\phantom{\{}\dspcl{\fly}{\penguin}{0,.05}\}\,. 
\eaq\]

This default theory should entail
that ``generally, penguins fly with a probability of at most 0.05''
(as properties of more specific classes should override 
inherited properties of less specific classes). Indeed, 
0-entailment yields the desired conclusion, while 
1-entailment reports an unsatisfiability (see item (7) in \mbox{Table \ref{EX-TAB-1}}). 

Moreover, $\BG_2$ should entail that ``generally, penguins 
have legs with a probability of at least 0.95'', since
penguins are exceptional birds w.r.t.\ to the ability 
of being able to fly, but not w.r.t.\ the property of having legs.  
However, 0-entailment provides only the interval $[0,1]$, and
1-entailment reports even an unsatisfiability (see item (5) in \mbox{Table \ref{EX-TAB-1}}). $\Box$
\ebs

The following example deals with the {\em drowning problem} (Benferhat {\em et al.}\ 1993). \nocite{BCDLP:lexi} 
\bbs\label{EX-3}
Let us consider the following probabilistic default theory $\BG_3\eqs(P_3,D_3)$:
\[\baq{l@{\ \ }l@{\ \ }l}
P_3 & = &\{\spcl{\bird}{\penguin}{1,1}\},\\[0.8ex]
D_3 & = &\{\dspcl{\fly}{\bird}{.9,.95},\,\dspcl{\fly}{\penguin}{0,.05},\\[0.8ex]
&&\phantom{\{}\dspcl{\easytosee}{\yellow}{.95,1}\}\,. 
\eaq\]

This default theory should entail that ``generally, yellow penguins are easy to see'', 
as the set of all yellow penguins is a nonexceptional subclass of the set of all yellow objects. 
But, 0-entailment gives only the interval $[0,1]$, and
1-entail\-ment reports an unsatisfiability (see item (8) in Table \ref{EX-TAB-1}). $\Box$
\ebs

The next example is taken from (Bacchus {\em et al.}\ 1996). \nocite{BGHK96}
\bbs\label{EX-4}
Let us consider the following probabilistic default theory $\BG_4\eqs(P_4,D_4)$:
\[\baq{l@{\ \ }l@{\ \ }l}
P_4 & = &\{\spcl{\bird}{\magpie}{1,1}\},\\[0.8ex]
D_4 & = &\{\dspcl{\chirps}{\bird}{.7,.8},\,\dspcl{\chirps}{\magpie}{0,.99}\}\,. 
\eaq\]

This default theory should entail ``generally, the probability that magpies chirp is between \mbox{0.7 and 0.8}'',
since we know more about birds w.r.t.\ the property of being able to chirp than about magpies.     
Indeed, both 0- and 1-entailment yield the desired conclusion (see item (9) in Table \ref{EX-TAB-1}). $\Box$
\ebs

The following example concerns {\em ambiguity preservation} (Benferhat {\em et al.}\ 1995). \nocite{benf-etal-98}
\bbs\label{EX-5}
Let us consider the following probabilistic default theory $\BG_5\eqs(P_5,D_5)$:
\[\baq{l@{\ \ }l@{\ \ }l}
P_5 & = &\{\spcl{\bird}{\penguin}{1,1}\},\\[0.8ex]
D_5 & = &\{\dspcl{\fly}{\winged}{.95,1},\,\dspcl{\fly}{\bird}{.95,1},\,\\[0.8ex]
&&\phantom{\{}\dspcl{\fly}{\penguin}{0,.05}\}\,. 
\eaq\]

Assume now that Oscar is a penguin with metal wings. As Oscar is a penguin, 
we should conclude that the probability that Oscar flies is at most 0.05. 
However, as Oscar has also metal wings, we should conclude that 
the probability that Oscar flies is at least 0.95.  
As argued in the literature on default reasoning (Benferhat {\em et al.}\ 1995), \nocite{benf-etal-98}
such ambiguities should be preserved. Indeed, 0-entailment
yields the desired interval $[0,1]$, while 1-entailment reports 
an unsatisfiability (see item (10) in Table \ref{EX-TAB-1}). $\Box$
\ebs

What about handling {\em purely probabilistic evidence}?
\bbs\label{EX-6}
Let us consider again the probabilistic default theory
$\BG_2$ of Example~\ref{EX-2}. Assume a first scenario in which 
our belief is ``the probability that Tweety is a bird is at least 0.9''
and ``the probability that Tweety is a penguin is at least 0.1'' and a second
scenario in which our belief is
``the probability that Tweety is a bird is at least 0.9''
and ``the probability that Tweety is a penguin is at least 0.9''.
What do we conclude about Tweety's ability to fly in these scenarios?

The notion of 0-entailment is undefined for such purely probabilistic evidence, 
whereas the notion of \mbox{1-entailment} yields the probability interval $[.86,.91]$ in the first scenario,
and reports an unsatisfiability in the second scenario (see items (11)--(12) in Table \ref{EX-TAB-1}). $\Box$
\ebs

Summarizing the results, 0-entailment is too weak, while 1-entailment is too strong. 
In detail, 0-entailment often yields the trivial interval $[0,1]$ 
and is even undefined for purely probabilistic evidence, while
1-entailment often reports unsatisfiabilities (in fact, in the most interesting scenarios,
as 1-entailment is actually {\em monotonic}).

Roughly speaking, our ideal notion of entailment for probabilistic knowledge bases under 
probabilistic default theories should lie somewhere between 0- and 1-entailment.   
One idea to obtain such a notion could be to strengthen 0-entailment 
by adding some inheritance mechanism. Another idea is to weaken 1-entailment
by handling unsatisfiabilities. In the rest of this paper, we will focus on the 
second idea. 

\section{Probabilistic Default Reasoning}\label{SEC-3}

In this section, we extend the classical notions of entailment in system $Z$ (Pearl 1990), \nocite{Pearl90} 
Lehmann's lexicographic entailment (1995), \nocite{Persp:94}
and Geffner's conditional entailment (Geffner 1992; Geffner \& Pearl 1992) 
\nocite{geffner:1992c,GeffnerPearl92} to conditional constraints. 

The main idea behind these extensions is to use the following two 
interpretations of defaults. 
As far as default rankings and priority orderings
are concerned, we interpret a default $(\psi\|\phi)[l,u]$ as 
``generally, \mbox{if $\phi$ is true}, then the probability of $\psi$ 
is between $l$ and  $u$''. Whereas, as far as notions
of entailment are concerned, we interpret
$(\psi\|\phi)[l,u]$ as ``the conditional probability of 
$\psi$ given $\phi$ is \mbox{between $l$ and  $u$}''.

\subsection{Preliminaries}

A probabilistic interpretation $\Pr$ {\em verifies} a default $(\hspace*{-1pt}\psi\|\phi\hspace*{-1pt})[l,\!u]$
iff $\Pr(\phi)\eqs 1$ and $\Pr\modelss(\psi|\phi)[l,u]$.
It {\em falsifies} a default $(\psi\|\phi)[l,u]$
iff $\Pr(\phi)\eqs 1$ and $\Pr\,{\not\models}\,(\psi|\phi)[l,u]$.
A set of defaults $D$ {\em tolerates} a default
$d$ {\em under} a set of strict conditional constraints $P$ iff 
$P\cup D$ has a model that verifies $d$. A set of defaults $D$ is
{\em under} $P$ {\em in conflict} with $d$ iff no 
model of $P\cup D$ verifies $d$. 

A {\em default ranking} $\sigma$ on $D$ maps 
each $d\ins D$ to a nonnegative integer. It is {\em admissible} with $\BG\eqs(P,D)$ iff each 
set of defaults $D'\subseteqs D$ that is under $P$ in conflict with 
some default $d\ins D$ contains a default $d'$ such that $\sigma(d')\lts\sigma(d)$. 
A probabilistic default theory $\BG\eqs (P,D)$ is {\em $\sigma$-consistent}
iff there exists a default ranking \mbox{on $D$} that is admissible with $\BG$. 
It is {\em $\sigma$-inconsistent} iff no such default ranking exists. 

A {\em probability ranking} $\kappa$ maps each 
probabilistic interpretation on $\isP$ to a member of
$\{0,1,\ldots\}\cup\{\infty\}$ such that $\kappa(\Pr)\eqs 0$ for at least one
interpretation $\Pr$. 
It is extended to all strict probabilistic formulas $F$ as
follows.  If $F$ is satisfiable, then $\kappa(F) =
\min\,\{\kappa(\Pr)\,|\,\Pr\models F\} $; otherwise,
$\kappa(F)=\infty$.  We say $\kappa$ is {\em admissible}
with $F$ iff $\kappa(\neg F)=\infty$. 
It is {\em admissible} with a default $(\psi\|\phi)[l,u]$ iff 
\[\hspace*{-3ex}\baq{l}\mbox{$\kappa((\phi|\top)[1,1])<\infty$ \ and}\\[0.8ex] 
\mbox{$\kappa((\phi|\top)[1,1]\wedges (\psi|\phi)[l,u])\lts
\kappa((\phi|\top)[1,1]\wedges \neg (\psi|\phi)[l,u])$\,.}\eaq\]
Roughly speaking, the intuition behind this definition is to interpret $(\psi\|\phi)[l,u]$ as
``generally, if $\phi$ is true, then
the probability of $\psi$ is between $l$ and  $u$''. 
A probability ranking $\kappa$ is {\em admissible} with a 
probabilistic default theory $\BG\eqs(P,D)$ iff 
$\kappa$ is admissible with all $F\ins P$ and all $d\ins D$. 

\begin{table*}[htb]
\renewcommand{\arraystretch}{1.3}
\caption{\label{EX-TAB-2}Examples of $z$-, lexicographically, and conditionally entailed tight intervals.}

\medskip
\begin{center}
\bt{|c||c|c|c||c|c|c|}\hline
 & $\BG$ & $\KB$ & $(\psi|\phi)$ & $\tazmodels$ & $\talmodels$ & $\tacmodels$\\\hline\hline
(1) & $\BG_1$ & $\spcl{\bird}{\top}{1,1}$ & $(\havelegs\,|\top)$ & $[.95,1]$ & $[.95,1]$ &  $[.95,1]$\\
(2) & $\BG_1$ & $\spcl{\bird\wedge\yellow}{\top}{1,1}$ & $(\havelegs\,|\top)$ & $[.95,1]$ & $[.95,1]$ & $[.95,1]$\\
(3) & $\BG_1$ & $\spcl{\penguin}{\top}{1,1}$ & $(\havelegs\,|\top)$  & $[.95,1]$ & $[.95,1]$ &  $[.95,1]$\\\hline
(4) & $\BG_2$ & $\spcl{\bird}{\top}{1,1}$ & $(\havelegs\,|\top)$  & $[.95,1]$ & $[.95,1]$ & $[.95,1]$\\
(5) & $\BG_2$ & $\spcl{\penguin}{\top}{1,1}$ & $(\havelegs\,|\top)$  & $\boldsymbol{[0,1]}$ & $[.95,1]$ & $[.95,1]$\\
(6) & $\BG_2$ & $\spcl{\bird}{\top}{1,1}$ & $(\fly\,|\top)$  & $[.9,.95]$ & $[.9,.95]$ & $[.9,.95]$\\
(7) & $\BG_2$ & $\spcl{\penguin}{\top}{1,1}$ & $(\fly\,|\top)$  & $[0,.05]$ & $[0,.05]$ & $[0,.05]$\\\hline
(8) & $\BG_3$ & $\spcl{\penguin\wedge \yellow}{\top}{1,1}$ & $(\easytosee\,|\top)$  &$\boldsymbol{[0,1]}$  & $[.95,1]$ &$[.95,1]$\\\hline
(9) & $\BG_4$ & $\spcl{\magpie}{\top}{1,1}$ & $(\chirps\,|\top)$  & $[.7,.8]$ & $[.7,.8]$ & $[.7,.8]$\\\hline
(10) & $\BG_5$ & $\spcl{\penguin\wedge \winged}{\top}{1,1}$ & $(\fly\,|\top)$ & $\boldsymbol{[0,.05]}$  & $\boldsymbol{[0,.05]}$ & $[0,1]$\\\hline
(11) & $\BG_2$ & $\spcl{\bird}{\top}{.9,1}\wedge\spcl{\penguin}{\top}{.1,1}$ & $(\fly\,|\top)$  & $[.86,.91]$ & $[.86,.91]$ & $[.86,.91]$\\
(12) & $\BG_2$ & $\spcl{\bird}{\top}{.9,1}\wedge\spcl{\penguin}{\top}{.9,1}$ & $(\fly\,|\top)$  & $[0,.15]$ & $[0,.15]$ & $[0,.15]$\\\hline
\et
\end{center}
\renewcommand{\arraystretch}{1}
\end{table*}

\subsection{System Z}

We now extend the notion of entailment in system $Z$ (Pearl 1990; Goldszmidt \& Pearl 1996) 
\nocite{Pearl90,goldszmidt-pearl:1996} to conditional constraints. 

In the sequel, let $\BG\eqs(P,D)$
be a $\sigma$-consistent probabilistic default theory. 
The notion of $z$-entailment is linked to an ordered
partition of $D$, a default ranking $z$, and a probability ranking $\kappa^z$.

We first define the $z$-partition of $D$. Let $(D_0,\ldots{},D_k)$ be the unique ordered partition of $D$ such that, for $i=0,\ldots{},k$, 
each $D_i$ is the set of all defaults in $D\setminus\bigcup\{D_j\,|\,0$$\,\le $$\,j$$\,< $$\,i\}$
that are tolerated under $P$ by $D\setminus\bigcup\{D_j\,|\,0$$\,\le $$\,j$$\,< $$\,i\}$
(note that we define $D\setminus\bigcup\{D_j\,|\,0\les j\lts i\}= D$ for $i\eqs 0$). 
We call this $(D_0,\ldots{},D_k)$ the {\em $z$-partition} of $D$. 
\bbs
The $z$-partition for the probabilistic default theory $\BG_2\eqs(P_2,D_2)$ 
of Example \ref{EX-2} is given as follows:
\[\baq{l}
(\{\dspcl{\havelegs}{\bird}{.95,1},
\dspcl{\fly}{\bird}{.9,.95}\},\,\\[0.8ex]
\phantom{(}\{\dspcl{\fly}{\penguin}{0,.05}\})\,. \ \ \Box
\eaq\]
\ebsw

We now define the default ranking $z$. For $j=0,\ldots{},k$, 
each $d$$\,\in$$\,D_j$ is assigned the value $j$ \mbox{under $z$}. 
The probability ranking $\kappa^z$ on all probabilistic interpretations $\Pr$ is then defined as follows: 
\[\baq{lll}\!\!\!\kappa^z(\Pr) & \!\! =\!\! & \begin{cases} \infty & \text{if $\Pr\not\models P$}\\[0.8ex]
0 & \text{if $\Pr\models P\cup D$}\\[0.8ex]
1+\max\limits_{d\in D\colon \Pr\not\models d}z(d) 
& \text{otherwise.}\cr
\end{cases}
\eaq\]

The following result shows that, in fact, $z$ is a default ranking that is 
admissible with $\BG$, and $\kappa^z$ is a probability ranking
that is admissible with $\BG$.
\ble\label{LEM-4-1} 
a) $z$ is a default ranking admissible with $\BG$.\\[0.5ex]
\noindent b) $\kappa^z$ is a probability ranking admissible with $\BG$.
\ele

We next define a preference relation on probabilistic interpretations.
For probabilistic interpretations $\Pr$ and $\Pr'$, 
we say $\Pr$ is {\em $z$-preferable} to $\Pr'$ iff $\kappa^z(\Pr)\lts \kappa^z(\Pr')$. 
\mbox{A model} $\Pr$ of a set of probabilistic formulas $\cF$ is a {\em $z$-minimal model} 
of $\cF$ iff no model of $\cF$ is $z$-preferable \mbox{to $\Pr$}.

We are now ready to define the notion of $z$-entailment as follows. 
A strict probabilistic formula $F$ is a {\em  $z$-con\-sequence} of 
$\KB$, denoted $\KB\zmodels F$,  
iff each $z$-minimal model of $P\cup\{\KB\}$ satisfies $F$. 
A strict conditional constraint $(\psi|\phi)[l,u]$ is a {\em tight $z$-consequence} 
of $\KB$, denoted $\KB\tzmodels (\psi|\phi)[l,u]$,
iff $l$ (resp., $u$) is the infimum (resp., supremum) of
$\Pr(\psi|\phi)$ subject to all $z$-minimal models $\Pr$ of $P\cup\{\KB\}$ with $\Pr(\phi)\gts 0$.

Coming back to Examples~\ref{EX-1}--\ref{EX-6}, it turns out that
the nonmonotonic properties of $z$-entailment differ from 
the ones of 0- and 1-entailment (see Table~\ref{EX-TAB-2}). 

In detail, in the given examples, $z$-entailment ignores irrelevant information, 
shows property inheritance to globally nonexceptional subclasses, and respects the principle of specificity. 
Moreover, it may also handle purely probabilistic evidence.
However, properties are still not inherited 
to more specific classes that are exceptional with respect to some other properties. 
Moreover, $z$-entailment still has the drowning problem 
and does not preserve ambiguities. 

The following examples illustrate how 
$z$-entailed tight intervals are determined.
\bbs
Given $\BG_2$ of Example \ref{EX-2},
we get: 
\[\spcl{\penguin}{\top}{1,1}\tzmodels\spcl{\havelegs}{\top}{0,1}\,\]
Here, the interval ``$[0,1]$'' comes from the tight logical consequence
$P_2\cup\{\dspcl{\fly}{\penguin}{0,.05},\spcl{\penguin}{\top}{1,1}\}$ $\tmodels\spcl{\havelegs}{\top}{0,1}$.  $\Box$
\ebs
\bbs
Given $\BG_5$ of Example \ref{EX-5},
we get:
\[\spcl{\penguin\wedge\winged}{\top}{1,1}\tzmodels\spcl{\fly}{\top}{0,.05}\,.\]
Here, the interval ``$[0,.05]$'' comes from the tight logical consequence
$P_5\cup\{\dspcl{\fly}{\penguin}{0,\!.05},\spcl{\penguin\wedge\winged\!}{\!\top}{1,1}\}%
\,{\tmodels}\, \spcl{\fly}{\top}{0,.05}$.  $\Box$
\ebs

\subsection{Lexicographic Entailment}

We now extend Lehmann's lexicographic entailment (Lehmann 1995) \nocite{Persp:94} to conditional constraints. 

In the sequel, let $\BG\eqs(P,D)$ be a $\sigma$-consistent probabilistic default theory. 
We now use the $z$-partition $(D_0,\ldots{},D_k)$ of $D$ to define a lexicographic
preference relation on probabilistic interpretations. 

For probabilistic interpretations $\Pr$ and $\Pr'$, we say
$\Pr$ is {\em lexicographically preferable} to $\Pr'$ 
iff there exists some $i\ins\{0,\ldots{},k\}$ such that $|\{d\ins D_i\,|\,\Pr\modelss
d\}|\gts|\{d\ins D_i\,|$ $\Pr'\modelss d\}|$ and $|\{d\ins D_j\,|\, \Pr\modelss
d\}| =|\{d\ins D_j\,|\,\Pr'\modelss d\}|$ for all $i\lts j\les k$.
A model $\Pr$ of a set of probabilistic formulas $\cF$ is a {\em lexicographically minimal model} 
of $\cF$ iff no model of $\cF$ is lexicographically preferable \mbox{to $\Pr$}.

We now define the notion of lexicographic entailment as follows. 
A strict probabilistic formula 
$F$ is a {\em  lexicographic consequence} of 
$\KB$, denoted $\KB\lmodels F$,  
iff each lexicographically minimal model of $P\cup \{\KB\}$ satisfies $F$.
A strict conditional constraint $(\psi|\phi)[l,u]$ is a {\em tight lexicographic consequence} 
of $\KB$, denoted $\KB\tlmodels (\psi|\phi)[l,u]$,
iff $l$ (resp., $u$) is the infimum (resp., supremum) of
$\Pr(\psi|\phi)$ subject to all lexicographically minimal  models $\Pr$ of $P\cup \{\KB\}$ with $\Pr(\phi)\gts 0$.

Coming back to Examples~\ref{EX-1}--\ref{EX-6}, it turns out that
lexicographic entailment has nicer nonmonotonic features  
than \mbox{$z$-entailment} (see Table~\ref{EX-TAB-2}). 

In detail, in the given examples, lexicographic entailment ignores irrelevant information, 
shows property inheritance to nonexceptional subclasses, and respects the principle of specificity. 
Moreover, it does not block
property inheritance, it does not have the drowning problem,
and it may also handle purely probabilistic evidence.
However, lexicographic entailment still does not preserve ambiguities. 

The following examples illustrate how 
lexicographically entailed tight intervals are determined.
\bbs
Given $\BG_2$ of Example \ref{EX-2},
we get:
\[\spcl{\penguin}{\top}{1,1}\tlmodels\spcl{\havelegs}{\top}{.95,1}\,.\]
Here, the interval ``$[.95,1]$'' comes from the tight logical consequence
$P_2\cup\{\dspcl{\havelegs}{\bird}{.95,1},\dspcl{\fly}{\penguin}{0,.05}$, 
$\spcl{\penguin}{\top}{1,1}\}\tmodels\spcl{\havelegs}{\top}{.95,1}$.  $\Box$
\ebs
\bbs
Given $\BG_5$ of Example \ref{EX-5},
we get: 
\[\spcl{\penguin\wedge\winged}{\top}{1,1}\tlmodels\spcl{\fly}{\top}{0,.05}\,.\]
Here, the interval ``$[0,.05]$'' comes from the tight logical consequence
$P_5\cup\{\dspcl{\fly}{\penguin}{0,\!.05},\spcl{\penguin\wedge\winged\!}{\!\top}{1,1}\}%
\,{\tmodels}\, \spcl{\fly}{\top}{0,.05}$.  $\Box$
\ebs

\subsection{Conditional Entailment}

We next extend Geffner's conditional entailment (Geffner 1992; Geffner \& Pearl 1992)
\nocite{geffner:1992c,GeffnerPearl92}
to conditional constraints. 

In the sequel, let $\BG\eqs(P,D)$ be a probabilistic default theory. 

We first define priority orderings on $D$ as follows.  
A {\em priority ordering} $\prec$ on $D$ is an irreflexive and transitive 
binary relation on $D$. We say $\prec$ is {\em admissible} 
with $\BG$ iff each set of defaults $D'\subseteqs D$ that is \mbox{under $P$} in conflict with 
some default $d\ins D$ contains a default $d'$ such that $d'\prec d$. 
We say $\BG$ is {\em $\prec$-consistent} iff 
there exists a priority ordering on $D$ that is admissible with $\BG$.
\bbs
Consider the probabilistic default theory $\BG_2\eqs(P_2,D_2)$ of Example \ref{EX-2}. 
A priority ordering $\prec$ on $D_2$ that is admissible with $\BG_2$
is given by $\dspcl{\fly}{\bird}{.9,.95}\prec \dspcl{\fly}{\penguin}{0,.05}$.  $\Box$
\ebs

The existence of an admissible default ranking implies the 
existence of an admissible priority ordering.
\ble\label{LEM-4-6}
If $\BG$ is $\sigma$-consistent, then $\BG$ is $\prec$-consistent.
\ele 

We next define a preference ordering on probabilistic interpretations as follows. 
Let $\Pr$ and $\Pr'$ be two probabilistic interpretations and let $\prec$ be a priority ordering on $D$. 
We say that $\Pr$ is {\em $\prec$-preferable} to $\Pr'$ iff 
$\{d\ins D\,|\,\Pr\,{\not\models}\,d\}\neq\{d\ins D\,|\,\Pr'\,{\not\models}\, d\}$ and
for each $d\ins D$ such that $\Pr\,{\not\models}\, d$ and $\Pr'\modelss d$, 
there exists some default $d'\ins D$ such that $d\prec d'$, $\Pr\modelss d'$, and
$\Pr'\,{\not\models}\, d'$. A model $\Pr$ of a set of probabilistic formulas $\cF$ 
\mbox{is a} {\em $\prec$-minimal model} 
of $\cF$ iff no model of $\cF$ is \mbox{$\prec$-preferable to $\Pr$}.
A model $\Pr$ of a set of probabilistic formulas $\cF$ is a 
{\em conditionally minimal model} of 
$\cF$ iff $\Pr$ is a {\em $\prec$-minimal model} of $\cF$ for some
priority ordering $\prec$ admissible with $\BG$.

We finally define the notion of conditional entailment. 
A strict probabilistic formula $F$ is a {\em  conditional consequence} of 
$\KB$, denoted $\KB\cmodels F$,  
iff each conditionally minimal model of \mbox{$P\cup \{\KB\}$} satisfies $F$. 
A strict conditional constraint $(\psi|\phi)[l,u]$ is a {\em tight conditional consequence} 
of $\KB$, denoted $\KB\tcmodels$ $(\psi|\phi)[l,u]$,
iff $l$ (resp., $u$) is the infimum (resp., supremum) of
$\Pr(\psi|\phi)$ subject to all conditionally minimal models $\Pr$ of $P\cup \{\KB\}$ 
with $\Pr(\phi)\gts 0$. 

Coming back to Examples~\ref{EX-1}--\ref{EX-6},
we see that among all introduced notions of entailment, 
conditional entailment is the one with the nicest 
nonmonotonic properties  (see Table~\ref{EX-TAB-2}). 

In detail, in the given examples, conditional entailment ignores irrelevant information, 
shows property inheritance to nonexceptional subclasses, and respects the principle of specificity. 
Moreover, it does not block
property inheritance, and it does not have the drowning problem.
Finally, conditional entailment preserves ambiguities
and may also handle purely probabilistic evidence.

The following examples illustrate how 
conditionally entailed tight intervals are determined.
\bbs
Given $\BG_2$ of Example \ref{EX-2},
we get: 
\[\spcl{\penguin}{\top}{1,1}\tcmodels\spcl{\havelegs}{\top}{.95,1}\,.\]
Here, the interval ``$[.95,1]$'' comes from the tight logical consequence
$P_2\cup\{\dspcl{\havelegs}{\bird}{.95,1},\dspcl{\fly}{\penguin}{0,.05}$, 
$\spcl{\penguin}{\top}{1,1}\}\tmodels\spcl{\havelegs}{\top}{.95,1}$.  $\Box$
\ebs
\bbs
Given $\BG_5$ of Example \ref{EX-5},
we get: 
\[\spcl{\penguin\wedge\winged}{\top}{1,1}\tcmodels\spcl{\fly}{\top}{0,1}\,.\]
Here, the interval ``$[0,1]$'' is the convex hull of the intervals ``$[0,.05]$'' and ``$[.95,1]$'',
which come from the tight logical consequences $P_5\cup\{\dspcl{\fly}{\penguin}{0,.05},
(\penguin\,\wedge$ $\winged\,|\top)[1,1]\}%
{\tmodels}\spcl{\fly}{\top}{0,.05}$ and 
$P_5{\cup}\{(\fly\,\|$ $\bird)[.95,1]$, $(\fly\,\|\,\winged)[.95,1]$,
$(\penguin\wedge\textit{metal\_}$ $\textit{wings}\,|\,\top)[1,1]\}%
\tmodels \spcl{\fly}{\top}{.95,1}$, respectively.   $\Box$
\ebs

\section{Relationship to Classical Formalisms}\label{SEC-REL}

We now analyze the relationship to classical default reasoning from conditional knowledge bases
and to classical probabilistic reasoning with conditional constraints. 

A {\em logical formula} is a probabilistic formula that contains only 
conditional constraints of the kind \mbox{$(\psi|\phi)[1\hspace*{-1pt},\!1]$\hspace*{-2pt}} or $(\psi\|\phi)[1,1]$.
A {\em strict logical formula} is a strict probabilistic formula that contains only 
strict conditional constraints of the form $(\psi|\phi)[1,1]$.
A {\em logical default theory} $T$ is a probabilistic default theory
that contains only logical formulas. 
A {\em logical knowledge base} $\KB$ is a strict logical formula.   

We use the operator $\gamma$ on logical formulas, sets of logical 
formulas, and logical default theories, which replaces each
strict conditional constraint $(\psi|\phi)[1,1]$ (resp., defeasible conditional constraint $(\psi\|\phi)[1,1]$)
by the classical implication $\phi\Ras\psi$ (resp., classical default $\phi\ras\psi$).  
Given a logical default theory $T$, we use $\oazmodels$ (resp., $\oalmodels$, $\oacmodels$)
to denote the classical notion of $z$-, (resp., lexicographic, conditional) 
entailment with respect to $\cl{T}$.         

The following result shows that the introduced notions of $z$-, lexicographic, and conditional entailment 
are generalizations of their classical counterparts.  
\bs\label{COR-4-3}
Let $\BG\eqs(P,D)$ be a logical default theory and let 
$\KB$ be a logical knowledge base. Then, for every semantics $s\ins\{z,{\it lex},{\it ce}\}$:\vspace*{0.1ex} 
\[\mbox{$\KB\asmodels (\psi|\top)[1,1]$ \ iff \ $\cl{\KB}\oasmodels \psi$.}\vspace*{0.5ex} \]
\es

The next result shows that, when the union of generic and concrete probabilistic knowledge 
is satisfiable, the notions of $z$-, lexicographic, and conditional entailment
coincide with the notion of 1-entailment.
\bs\label{THE-4-3}
Let $\BG\eqs(P,D)$ be a probabilistic default theory
and let $\KB$ be a probabilistic knowledge base such that $P\cups D\cups \{\KB\}$ is satisfiable.  
Then, for every semantics $s\ins\{z,{\it lex},{\it ce}\}$:
\be
\item $\KB\asmodels F$ \ iff \ $P\cups D\cups \{\KB\}\models F$.
\item $\KB{\tasmodels}(\psi|\phi)[l,u]$ iff $P{\cup} D{\cup} \{\KB\}{\tmodels}(\psi|\phi)[l,u]$.
\ee
\es

\section{Summary and Outlook}\label{SEC-4}

We presented the notions of $z$-entailment, lexicographic entailment, 
and conditional entailment for conditional constraints, which combine the classical notions of 
entailment in system $Z$, 
Lehmann's lexicographic entailment, and Geffner's conditional entailment with the
classical notion of logical entailment for conditional constraints.
We showed that the introduced notions for probabilistic default reasoning 
with conditional constraints have similar properties like their classical counterparts. Moreover, 
they properly extend both their classical counterparts
and the classical notion of logical entailment for conditional constraints.

An interesting topic of future research is to extend other formalisms
for classical default reasoning to the probabilistic framework of conditional constraints.  

\section{Acknowledgments}

I am very grateful to the referees for their useful comments. 

This work was supported by a DFG grant and the Austrian Science Fund Project N Z29-INF.

\nocite{eite-luka-00,Lukas00-1}
\nocite{deF74}
\nocite{Carnap50}
\nocite{FHM92}


\begin{thebibliography}{}

\bibitem[\protect\citeauthoryear{Adams}{1975}]{Ada75}
Adams, E.~W.
\newblock 1975.
\newblock {\em The Logic of Conditionals}, volume~86 of {\em Synthese Library}.
\newblock Dordrecht, Netherlands: D. Reidel.

\bibitem[\protect\citeauthoryear{Amarger, Dubois, \& Prade}{1991}]{ADP91a}
Amarger, S.; Dubois, D.; and Prade, H.
\newblock 1991.
\newblock Constraint propagation with imprecise conditional probabilities.
\newblock In {\em Proceedings UAI-91},  26--34.
\newblock Morgan Kaufmann.

\bibitem[\protect\citeauthoryear{Bacchus \bgroup \em et al.\egroup
  }{1996}]{BGHK96}
Bacchus, F.; Grove, A.; Halpern, J.; and Koller, D.
\newblock 1996.
\newblock From statistical knowledge bases to degrees of beliefs.
\newblock {\em Artif.\ Intell.} 87:75--143.

\bibitem[\protect\citeauthoryear{Benferhat \bgroup \em et al.\egroup
  }{1993}]{BCDLP:lexi}
Benferhat, S.; Cayrol, C.; Dubois, D.; Lang, J.; and Prade, H.
\newblock 1993.
\newblock Inconsistency management and prioritized syntax-based entailment.
\newblock In {\em Proceedings IJCAI-93},  640--645.
\newblock Morgan Kaufmann.

\bibitem[\protect\citeauthoryear{Benferhat, Dubois, \& Prade}{1992}]{BDP92}
Benferhat, S.; Dubois, D.; and Prade, H.
\newblock 1992.
\newblock Representing default rules in possibilistic logic.
\newblock In {\em Proceedings KR-92},  673--684.
\newblock Morgan Kaufmann.

\bibitem[\protect\citeauthoryear{Benferhat, Dubois, \& Prade}{1997}]{BDH97}
Benferhat, S.; Dubois, D.; and Prade, H.
\newblock 1997.
\newblock Nonmonotonic reasoning, conditional objects and possibility theory.
\newblock {\em Artif.\ Intell.} 92(1--2):259--276.

\bibitem[\protect\citeauthoryear{Benferhat, Saffiotti, \&
  Smets}{1995}]{benf-etal-98}
Benferhat, S.; Saffiotti, A.; and Smets, P.
\newblock 1995.
\newblock Belief functions and default reasoning.
\newblock In {\em Proceedings UAI-95},  19--26.
\newblock Morgan Kaufmann.

\bibitem[\protect\citeauthoryear{Carnap}{1950}]{Carnap50}
Carnap, R.
\newblock 1950.
\newblock {\em Logical Foundations of Probability}.
\newblock Chicago: University of Chicago Press.

\bibitem[\protect\citeauthoryear{de Finetti}{1974}]{deF74}
de~Finetti, B.
\newblock 1974.
\newblock {\em Theory of Probability}.
\newblock New York: Wiley.

\bibitem[\protect\citeauthoryear{Dubois \& Prade}{1988}]{DP88}
Dubois, D., and Prade, H.
\newblock 1988.
\newblock On fuzzy syllogisms.
\newblock {\em Computational Intelligence} 4(2):171--179.

\bibitem[\protect\citeauthoryear{Dubois \& Prade}{1991}]{DP91}
Dubois, D., and Prade, H.
\newblock 1991.
\newblock Possibilistic logic, preferential models, non-monotonicity and
  related issues.
\newblock In {\em Proceedings IJCAI-91},  419--424.
\newblock Morgan Kaufmann.

\bibitem[\protect\citeauthoryear{Dubois \& Prade}{1994}]{DP94}
Dubois, D., and Prade, H.
\newblock 1994.
\newblock Conditional objects as non-monotonic consequence relationships.
\newblock {\em {IEEE} Trans.\ Syst.\ Man Cybern.} 24:1724--1740.

\bibitem[\protect\citeauthoryear{Dubois \bgroup \em et al.\egroup
  }{1993}]{DPGM93}
Dubois, D.; Prade, H.; Godo, L.; and de~M{\`a}ntaras, R.~L.
\newblock 1993.
\newblock Qualitative reasoning with imprecise probabilities.
\newblock {\em Journal of Intelligent Information Systems} 2:319--363.

\bibitem[\protect\citeauthoryear{Dubois, Prade, \& Touscas}{1990}]{DPT90}
Dubois, D.; Prade, H.; and Touscas, J.-M.
\newblock 1990.
\newblock Inference with imprecise numerical quantifiers.
\newblock In Ras, Z.~W., and Zemankova, M., eds., {\em Intelligent Systems}.
  Ellis Horwood.
\newblock chapter~3,  53--72.

\bibitem[\protect\citeauthoryear{Eiter \& Lukasiewicz}{2000}]{eite-luka-00}
Eiter, T., and Lukasiewicz, T.
\newblock 2000.
\newblock Complexity results for default reasoning from conditional knowledge
  bases.
\newblock In {\em Proceedings KR-2000}.
\newblock Morgan Kaufmann.
\newblock To appear.

\bibitem[\protect\citeauthoryear{Fagin, Halpern, \& Megiddo}{1990}]{FHM92}
Fagin, R.; Halpern, J.~Y.; and Megiddo, N.
\newblock 1990.
\newblock A logic for reasoning about probabilities.
\newblock {\em Inf.\ Comput.} 87:78--128.

\bibitem[\protect\citeauthoryear{Friedman \& Halpern}{2000}]{FH99}
Friedman, N., and Halpern, J.~Y.
\newblock 2000.
\newblock Plausibility measures and default reasoning.
\newblock {\em J.\ ACM}.
\newblock To appear.

\bibitem[\protect\citeauthoryear{Frisch \& Haddawy}{1994}]{FH94+}
Frisch, A.~M., and Haddawy, P.
\newblock 1994.
\newblock Anytime deduction for probabilistic logic.
\newblock {\em Artif.\ Intell.} 69:93--122.

\bibitem[\protect\citeauthoryear{Geffner \& Pearl}{1992}]{GeffnerPearl92}
Geffner, H., and Pearl, J.
\newblock 1992.
\newblock Conditional entailment: Bridging two approaches to default reasoning.
\newblock {\em Artif.\ Intell.} 53(2--3):209--244.

\bibitem[\protect\citeauthoryear{Geffner}{1992}]{geffner:1992c}
Geffner, H.
\newblock 1992.
\newblock {\em Default Reasoning: Causal and Conditional Theories}.
\newblock Cambridge, MA: {MIT} Press.

\bibitem[\protect\citeauthoryear{Goldszmidt \& Pearl}{1992}]{GP92}
Goldszmidt, M., and Pearl, J.
\newblock 1992.
\newblock Rank-based systems: {A} simple approach to belief revision, belief
  update and reasoning about evidence and actions.
\newblock In {\em Proceedings KR-92},  661--672.
\newblock Morgan Kaufmann.

\bibitem[\protect\citeauthoryear{Goldszmidt \&
  Pearl}{1996}]{goldszmidt-pearl:1996}
Goldszmidt, M., and Pearl, J.
\newblock 1996.
\newblock Qualitative probabilities for default reasoning, belief revision, and
  causal modeling.
\newblock {\em Artif.\ Intell.} 84(1--2):57--112.

\bibitem[\protect\citeauthoryear{Goldszmidt, Morris, \& Pearl}{1993}]{GPM93}
Goldszmidt, M.; Morris, P.; and Pearl, J.
\newblock 1993.
\newblock A maximum entropy approach to nonmonotonic reasoning.
\newblock {\em {IEEE} Trans.\ Pattern Anal.\ Mach.\ Intell.} 15(3):220--232.

\bibitem[\protect\citeauthoryear{Grove, Halpern, \& Koller}{1994}]{GHK94}
Grove, A.; Halpern, J.; and Koller, D.
\newblock 1994.
\newblock Random worlds and maximum entropy.
\newblock {\em J.\ Artif.\ Intell.\ Res.} 2:33--88.

\bibitem[\protect\citeauthoryear{Heinsohn}{1994}]{Hein94}
Heinsohn, J.
\newblock 1994.
\newblock Probabilistic description logics.
\newblock In {\em Proceedings UAI-94}.
\newblock Morgan Kaufmann.

\bibitem[\protect\citeauthoryear{Jaumard, Hansen, \& de
  Arag{\~{a}}o}{1991}]{jahapo91}
Jaumard, B.; Hansen, P.; and de~Arag{\~{a}}o, M.~P.
\newblock 1991.
\newblock Column generation methods for probabilistic logic.
\newblock {\em ORSA J.\ Comput.} 3:135--147.

\bibitem[\protect\citeauthoryear{Kraus, Lehmann, \& Magidor}{1990}]{KLM90}
Kraus, S.; Lehmann, D.; and Magidor, M.
\newblock 1990.
\newblock Nonmonotonic reasoning, preferential models and cumulative logics.
\newblock {\em Artif.\ Intell.} 14(1):167--207.

\bibitem[\protect\citeauthoryear{{Kyburg,~Jr.}}{1974}]{Kyb74}
{Kyburg,~Jr.}, H.~E.
\newblock 1974.
\newblock {\em The Logical Foundations of Statistical Inference}.
\newblock Dordrecht, Netherlands: D. Reidel.

\bibitem[\protect\citeauthoryear{{Kyburg,~Jr.}}{1983}]{Kyb83}
{Kyburg,~Jr.}, H.~E.
\newblock 1983.
\newblock The reference class.
\newblock {\em Philos.\ Sci.} 50:374--397.

\bibitem[\protect\citeauthoryear{Lamarre}{1992}]{Lamarre92}
Lamarre, P.
\newblock 1992.
\newblock A promenade from monotonicity to non-monotonicity following a theorem
  prover.
\newblock In {\em Proceedings KR-92},  572--580.
\newblock Morgan Kaufmann.

\bibitem[\protect\citeauthoryear{Lehmann \&
  Magidor}{1992}]{lehmann_d-magidor:1992b1}
Lehmann, D., and Magidor, M.
\newblock 1992.
\newblock What does a conditional knowledge base entail?
\newblock {\em Artif.\ Intell.} 55(1):1--60.

\bibitem[\protect\citeauthoryear{Lehmann}{1989}]{Lehmann89}
Lehmann, D.
\newblock 1989.
\newblock What does a conditional knowledge base entail?
\newblock In {\em Proceedings KR-89},  212--222.
\newblock Morgan Kaufmann.

\bibitem[\protect\citeauthoryear{Lehmann}{1995}]{Persp:94}
Lehmann, D.
\newblock 1995.
\newblock Another perspective on default reasoning.
\newblock {\em Ann.\ Math.\ Artif.\ Intell.} 15(1):61--82.

\bibitem[\protect\citeauthoryear{Lukasiewicz}{1998}]{Luk98c}
Lukasiewicz, T.
\newblock 1998.
\newblock Probabilistic logic programming.
\newblock In {\em Proceedings ECAI-98},  388--392.
\newblock J.\ Wiley {\&} Sons.

\bibitem[\protect\citeauthoryear{Lukasiewicz}{1999a}]{Luk99b}
Lukasiewicz, T.
\newblock 1999a.
\newblock Local probabilistic deduction from taxonomic and probabilistic
  knowledge-bases over conjunctive events.
\newblock {\em Int.\ J.\ Approx.\ Reas.} 21(1):23--61.

\bibitem[\protect\citeauthoryear{Lukasiewicz}{1999b}]{Luk99a}
Lukasiewicz, T.
\newblock 1999b.
\newblock Probabilistic deduction with conditional constraints over basic
  events.
\newblock {\em J.\ Artif.\ Intell.\ Res.} 10:199--241.

\bibitem[\protect\citeauthoryear{Lukasiewicz}{2000}]{Lukas00-1}
Lukasiewicz, T.
\newblock 2000.
\newblock Probabilistic default reasoning with strict and defeasible
  conditional constraints.
\newblock Technical Report INFSYS RR-1843-00-02, Institut f{\"u}r
  Informationssysteme, Technische Universit{\"a}t Wien.

\bibitem[\protect\citeauthoryear{Luo \bgroup \em et al.\egroup
  }{1996}]{LYLWP96}
Luo, C.; Yu, C.; Lobo, J.; Wang, G.; and Pham, T.
\newblock 1996.
\newblock Computation of best bounds of probabilities from uncertain data.
\newblock {\em Computational Intelligence} 12(4):541--566.

\bibitem[\protect\citeauthoryear{Pearl}{1989}]{Pearl89}
Pearl, J.
\newblock 1989.
\newblock Probabilistic semantics for nonmontonic reasoning: {A} survey.
\newblock In {\em Proceedings KR-89},  505--516.
\newblock Morgan Kaufmann.

\bibitem[\protect\citeauthoryear{Pearl}{1990}]{Pearl90}
Pearl, J.
\newblock 1990.
\newblock System {Z}: {A} natural ordering of defaults with tractable
  applications to default reasoning.
\newblock In {\em Proceedings TARK-90},  121--135.
\newblock Morgan Kaufmann.

\bibitem[\protect\citeauthoryear{Pollock}{1990}]{Pollock90}
Pollock, J.~L.
\newblock 1990.
\newblock {\em Nomic Probabilities and the Foundations of Induction}.
\newblock Oxford: Oxford University Press.

\bibitem[\protect\citeauthoryear{Reichenbach}{1949}]{Rei49}
Reichenbach, H.
\newblock 1949.
\newblock {\em Theory of Probability}.
\newblock Berkeley, CA: University of California Press.

\bibitem[\protect\citeauthoryear{Shoham}{1987}]{Shoham87}
Shoham, Y.
\newblock 1987.
\newblock A semantical approach to nonmonotonic logics.
\newblock In {\em Proceedings of the 2nd IEEE Symposium on Logic in Computer
  Science},  275--279.

\bibitem[\protect\citeauthoryear{Spohn}{1988}]{Spohn88}
Spohn, W.
\newblock 1988.
\newblock Ordinal conditional functions: {A} dynamic theory of epistemic
  states.
\newblock In Harper, W., and Skyrms, B., eds., {\em Causation in Decision,
  Belief Change, and Statistics}, volume~2. Dordrecht, Netherlands: Reidel.
\newblock  105--134.

\bibitem[\protect\citeauthoryear{Th{\"o}ne, G{\"u}ntzer, \&
  Kie{\ss{}}ling}{1992}]{TGK92}
Th{\"o}ne, H.; G{\"u}ntzer, U.; and Kie{\ss{}}ling, W.
\newblock 1992.
\newblock Towards precision of probabilistic bounds propagation.
\newblock In {\em Proceedings UAI-92},  315--322.
\newblock Morgan Kaufmann.

\end{thebibliography}

\end{document}